\documentclass[sigconf]{acmart}

\usepackage{booktabs} % For formal tables
\usepackage{multirow}
\usepackage{graphicx}
\usepackage{tabularx}
\usepackage{amsmath}
\usepackage{algorithmic} 
\usepackage{enumitem}
\usepackage{balance}
\setcopyright{none}
\acmConference{MM'17}{October 23--27, 2017}{Mountain View, CA, USA.} 
\acmYear{2017}
\copyrightyear{2017}

\acmPrice{15.00}

\begin{document}

\title{Fine-grained Discriminative Localization \\via Saliency-guided Faster R-CNN}
%\titlenote{Produces the permission block, and
%  copyright information}
%\subtitle{Extended Abstract}
%\subtitlenote{The full version of the author's guide is available as
%  \texttt{acmart.pdf} document}

\author{Xiangteng He, Yuxin Peng* and Junjie Zhao}
\affiliation{%
  \institution{Institute of Computer Science and Technology, Peking University}
  %\streetaddress{P.O. Box 1212}
  \city{Beijing} 
  \country{China} 
  \postcode{100871}
}
\email{pengyuxin@pku.edu.cn}

%\author{Julius P.~Kumquat}
%\affiliation{\institution{The Kumquat Consortium}}
%\email{jpkumquat@consortium.net}

\renewcommand{\shortauthors}{He et al.}

\begin{abstract}
Discriminative localization is essential for fine-grained image classification task, which devotes to recognizing hundreds of subcategories in the same basic-level category. Reflecting on discriminative regions of objects, key differences among different subcategories are subtle and local. 
%Most existing discriminative localization based methods for fine-grained image classification can generally be divided in two groups: (1) Weakly supervised methods, which use only image-level subcategory label, without using object or parts annotations. (2) End-to-end methods, which simultaneously localize discriminative regions and encode the discriminative features, improving the classification speed. 
%Despite achieving promising results, these methods have two limitations compared with each other: (1) Weakly supervised methods cause the testing \emph{time consuming}, due to the separation of discriminative localization and feature encoding. (2) End-to-end methods cause the labeling is heavily \emph{labor consuming}, due to the need of object or part annotations for training. 
Existing methods generally adopt a two-stage learning framework: \emph{The first stage} is to localize the discriminative regions of objects, and \emph{the second} is to encode the discriminative features for training classifiers. 
However, these methods generally have two limitations: 
(1) \emph{Separation} of the two-stage learning is \emph{time-consuming}. (2) \emph{Dependence} on object and parts annotations for discriminative localization learning leads to heavily \emph{labor-consuming} labeling. 
It is highly challenging to address these two important limitations \emph{simultaneously}. Existing methods only focus on one of them.
%Most existing methods for fine-grained image classification generally adopt a two-stage learning framework: \emph{The first learning stage} is to localize the discriminative regions of objects, and \emph{the second learning stage} is to encode the discriminative features for training classifiers. 
%Despite achieving promising results, these methods have two limitations: 
%(1) \emph{Separation} of these two-stage learning causes the process of classification is highly \emph{time consuming}. (2) \emph{Dependence} on the object and parts annotations for learning discriminative localization causes the labeling is heavily \emph{labor consuming}. 
Therefore, this paper proposes \emph{the discriminative localization approach via saliency-guided Faster R-CNN} to address the above two limitations at the same time, and our main novelties and advantages are: (1) \emph{End-to-end network} based on Faster R-CNN is designed to \emph{simultaneously} localize discriminative regions and encode discriminative features, which accelerates classification speed. (2) \emph{Saliency-guided localization learning} is proposed to localize the discriminative region automatically, avoiding labor-consuming labeling. Both are jointly employed to simultaneously accelerate classification speed and eliminate dependence on object and parts annotations.  
Comparing with the state-of-the-art methods on the widely-used CUB-200-2011 dataset, our approach achieves both the best classification accuracy and efficiency.
\end{abstract}


\maketitle

\section{Introduction}
\begin{figure}[!t]
    \begin{center}\includegraphics[width=1\linewidth]{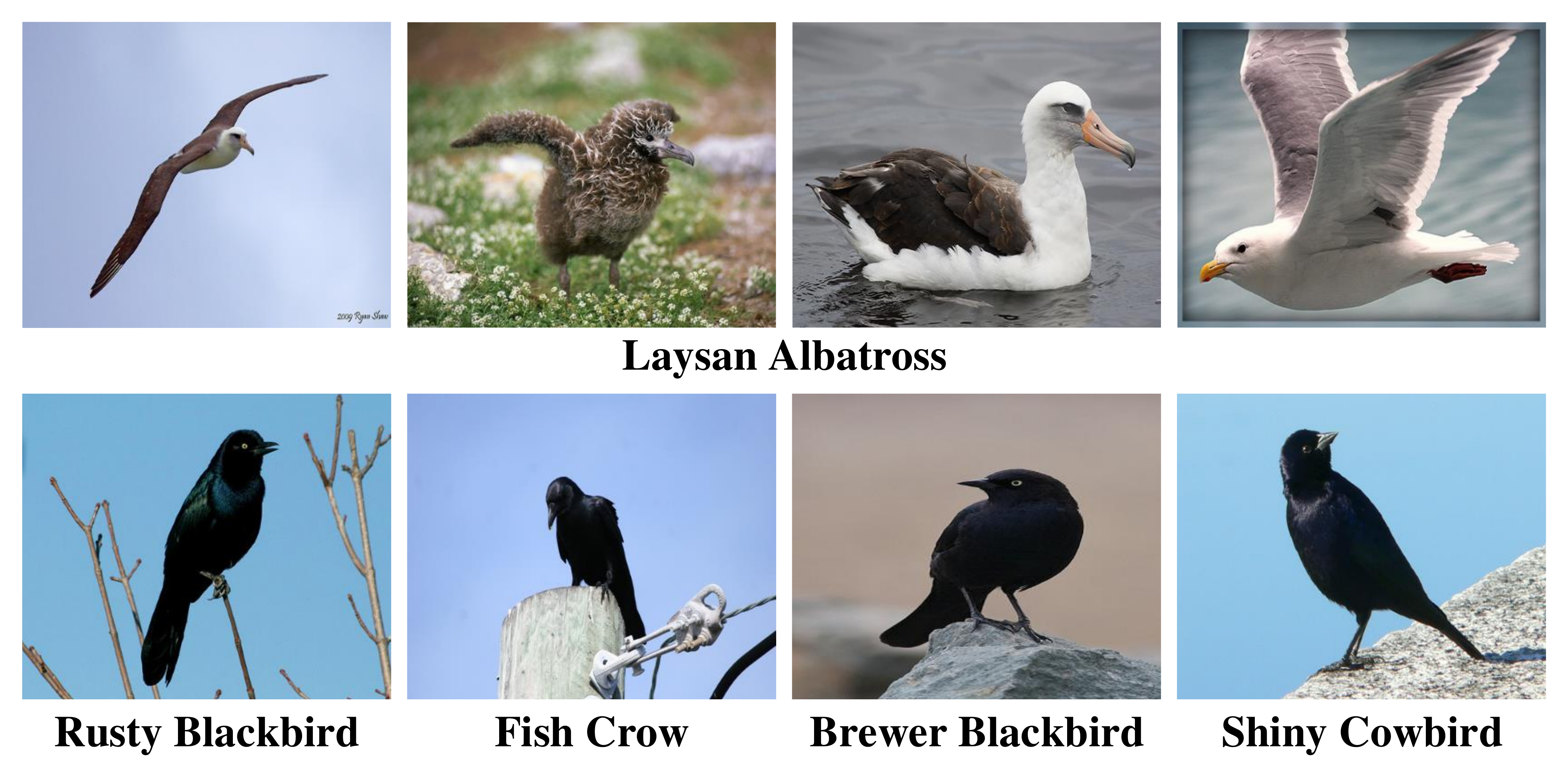}
    \caption{Examples of CUB-200-2011 dataset \cite{wah2011caltech}. First row shows large variance in the same subcategory, and second row shows small variance among different subcategories.}
    \label{interintra}
    \end{center}
\end{figure}

Fine-grained image classification is a highly challenging task due to large variance in the same subcategory and small variance among different subcategories, as shown in
Figure \ref{interintra}, which is to recognize hundreds of subcategories belonging to the same basic-level category. These subcategories look similar in global appearances, but have subtle differences at discriminative regions of objects, which are crucial for classification. Therefore, most researchers focus on localizing discriminative regions of objects to promote the performance of fine-grained image classification. 

Most existing methods \cite{zhang2014part,zhang2014fused,krause2015fine,xiao2015application,zhang2016picking,simon2015neural,he2017spatial} generally follow a two-stage learning framework: The first learning stage is to localize discriminative regions of objects, and the second is to encode the discriminative features for training classifiers. 
Girshick et al. \cite{girshick2014rich} propose a simple and scalable detection algorithm, called R-CNN. It generates thousands of region proposals for each image via bottom-up process \cite{uijlings2013selective} first. And then extracts features of objects via convolutional neural network (CNN) to train an object detector for each class, which is used to discriminate the probabilities of the region proposals being objects. This framework is widely used in fine-grained classification. 
Zhang et al. \cite{zhang2014part} utilize R-CNN with geometric constraints to detect object and its parts first, and then extract features for the object and its parts, finally train a one-versus-all linear SVM for classification. It needs both object and parts annotations. 
Krause et al. \cite{krause2015fine} adopt the box constraint of Part-based R-CNN \cite{zhang2014part} to train part detectors with only object annotation. 
These methods generally have two limitations: (1) Separation of the two-stage learning is time-consuming. (2) Dependence on object and parts annotations for discriminative localization learning leads to heavily labor-consuming labeling. It is highly challenging to address these two limitations simultaneously. Existing works only focus on one of them.

\begin{figure*}[!ht]
    \begin{center}\includegraphics[width=1\linewidth]{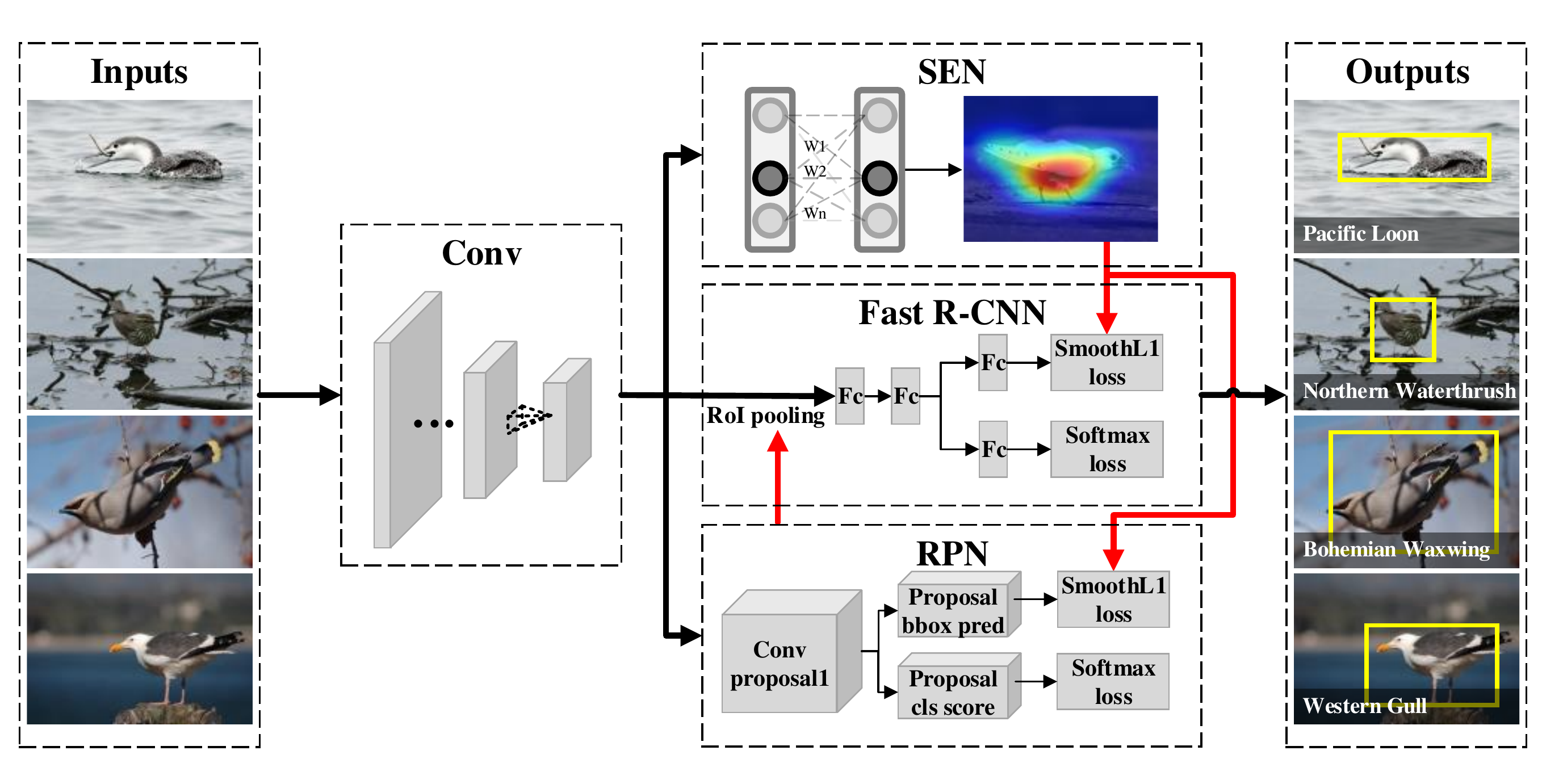}
    \caption{An overview of our Saliency-guided Faster R-CNN approach. Saliency extraction network (SEN) extracts the saliency information to provide the bounding box for training region proposal network (RPN)  and Fast R-CNN, RPN produces the region proposal to accelerate the process of proposal generation, and Fast R-CNN learns to localize the discriminative region. The outputs show the predicted discriminative regions and subcategories.}
    \label{framework}
    \end{center}
\end{figure*}

For addressing the first limitation, researchers focus on the end-to-end network. 
%Lin et al. \cite{lin2015bilinear} propose a bilinear CNN model, which jointly combining two CNNs, each of which is adopted as a feature extractor. The extracted features at each location of image are multiplied by outer product processing, and then pooled to generate an image descriptor. Finally, SVM is conducted for final prediction. It is faster than Part-based R-CNN, but could not localize the object and parts to tell what distinguish this from other subcategories.
Zhang et al. \cite{zhang2016spda} propose a Part-Stacked CNN architecture, which first utilizes a fully convolutional network to localize parts of object, and then adopts a two-stream classification network to encode object-level and part-level features simultaneously. It is over two order of magnitude faster than Part-based R-CNN \cite{zhang2014part}, but relies heavily on object and parts annotations that are \emph{labor consuming}. 
%Huang et al. \cite{huang2016part} 

For addressing the second limitation, researchers focus on how to localize the discriminative regions under the weakly supervised setting, which means neither object nor parts annotations are used in training or testing phase. 
Xiao et al. \cite{xiao2015application} propose a two-level attention model: object-level attention is to select relevant region proposals to a certain object, and part-level attention is to localize discriminative parts of object. 
It is the first work to classify fine-grained images without using object or parts annotations in both training and testing phase, but still achieves promising results \cite{zhang2016weakly}. 
Simon and Rodner \cite{simon2015neural} propose a constellation model to localize parts of object, leveraging CNN to find the constellations of neural activation patterns. A part model is estimated by selecting part detectors via constellation model. And then the part model is used to extract features for classification. Zhang et al. \cite{zhang2016picking} incorporate deep convolutional filters for both parts selection and description. He and Peng \cite{he2017spatial} integrate two spatial constraints for improving the performance of parts selection.
These methods rarely depend on object or parts annotations, but their classification speeds are \emph{time consuming} due to the separation of localization and encoding. 

Different from them, this paper proposes a discriminative localization approach via saliency-guided Faster R-CNN, which is the first attempt based on discriminative localization to simultaneously accelerate classification speed and eliminate dependence on object and parts annotations. Its main novelties and contributions can be summarized as follows:

\begin{itemize}[leftmargin=1.5em]
\item
\emph{\textbf{End-to-end network.}} \ Most existing discriminative localization based methods \cite{xiao2015application,simon2015neural,zhang2016picking} generally  localize discriminative regions first, and then encode discriminative features. The separated processes cause highly \emph{time-consuming} classification. 
For addressing this important problem, we propose an \emph{end-to-end network} based on Faster R-CNN to \emph{accelerate} the classification speed by simultaneously localizing discriminative regions and encoding discriminative features. Localization exploits discriminative regions with subtle but distinguishing features from other subcategories, and encoding generates representative descriptions. They have synergistic effect with each other, which further improves the classification performance.

\item
\emph{\textbf{Saliency-guided localization learning.}} \ Existing methods as \cite{huang2016part} combine localization and encoding to accelerate classification speed. However, localization learning relies heavily on object or parts annotations, which is \emph{labor consuming}.
For addressing this important problem, we propose a \emph{saliency-guided localization learning} approach, which \emph{eliminates the heavy dependence on object and parts annotations} by localizing the discriminative regions automatically. We adopt a neural network with global average pooling (GAP) layer, which is called saliency extraction network (SEN), to extract the saliency information for each image. And then share convolutional weights between SEN and Faster R-CNN to transfer knowledge of discriminative features. This takes the advantages of both SEN and Faster R-CNN to boost the discriminative localization and avoid the labor-consuming labeling simultaneously.
\end{itemize}

The rest of this paper is organized as follows: 
%Section \uppercase\expandafter{\romannumeral2} briefly reviews related works on fine-grained image classification. 
Section 2 presents our approach in detail, and Section 3 introduces the experiments as well as the results analyses. Finally Section 4 concludes this paper.

\section{Saliency-guided Faster R-CNN}
We propose a discriminative localization approach via saliency-guided Faster R-CNN without using object or parts annotations. Saliency-guided Faster R-CNN is an end-to-end network to localize discriminative regions and encode discriminative features simultaneously, which not only achieves a notable classification performance but also accelerates classification speed. It consists of two components: saliency extraction network (SEN) and Faster R-CNN. 
SEN extracts saliency information of each image for generating the bounding box which is used to guide the discriminative localization learning of Faster R-CNN. They are two localization learning stages, and their jointly learning further achieves better performance.
% It is a process of learning discriminative localization by learned discriminative region
An overview of our approach is shown as Figure \ref{framework}.

\subsection{Weakly supervised Faster R-CNN}
\begin{figure}[!t]
    \begin{center}\includegraphics[width=1\linewidth]{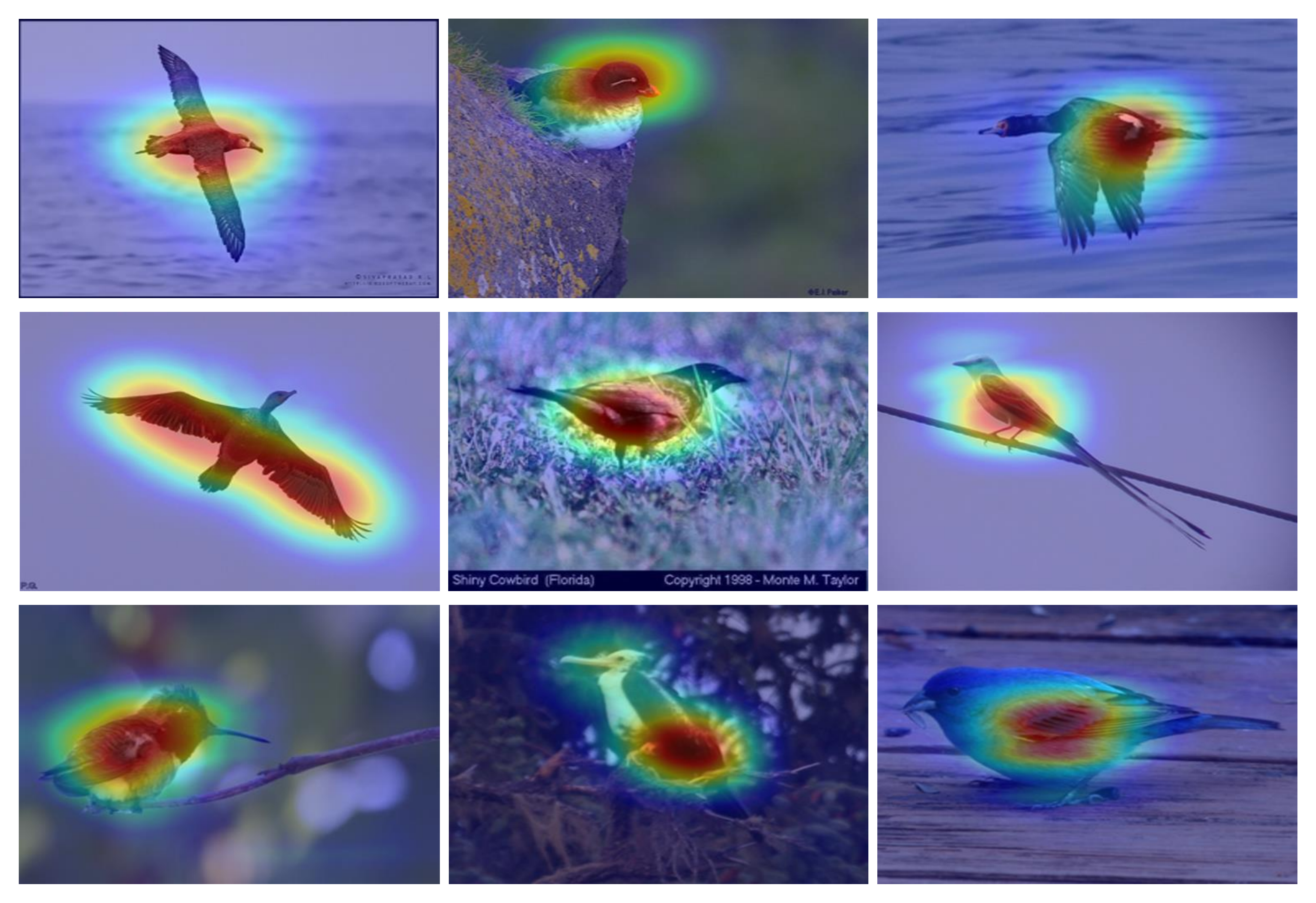}
    \caption{Some examples of saliency maps extracted by SEN in our Saliency-guided Faster R-CNN approach.}
    \label{saliencymap}
    \end{center}
\end{figure}

We propose a weakly supervised Faster R-CNN to accelerate classification speed and achieve promising results simultaneously without using object or parts annotations. 
A saliency extraction network (SEN) is proposed to generate bounding box information for Faster R-CNN first. It takes a resized image as an input and outputs a saliency map for generating the bounding box of discriminative region. We follow the work of Zhou et al. \cite{zhou2016cvpr} to model this process by utilizing global average pooling (GAP) to produce the saliency map. 
We sum the feature maps of last convolutional layer with weights to generate the saliency map for each image. Figure \ref{saliencymap} shows some examples of saliency maps obtained by our approach. Finally we perform binarization operation on the saliency map with a adaptive threshold, which is obtained via OTSU algorithm \cite{otsu1979threshold}, and take the bounding box that covers the largest connected area as the discriminative region of object.
%Through SEN, we obtain the saliency map $M_c$ of an image for subcategory $c$ to localize the discriminative region. And then we perform binarization and connected region extraction on the saliency maps to get the bounding box of localized region.
For a given image $I$, the value of spatial location $(x,y)$ in saliency map for subcategory $c$ is defined as follows:
\begin{gather}
M_c(x,y) = \sum \limits_u w_u^c f_u(x,y)
\end{gather}
where $M_c(x,y)$ directly indicates the importance of activation at spatial location $(x,y)$ leading to the classification of an image to subcategory $c$, $f_u(x,y)$ denotes the activation of neuron $u$ in the last convolutional layer at spatial location $(x,y)$, and $w_u^c$ denotes the weight that corresponding to subcategory $c$ for neuron $u$. 
Instead of using the image-level subcategory labels, we use the predicted result as the subcategory $c$.

Faster R-CNN \cite{ren2015faster} is proposed to accelerate the process of detection as well as achieve promising detection performance. However, the training phase needs ground truth bounding boxes of objects for supervised learning, which is heavily labor consuming. In this paper, we propose weakly supervised Faster R-CNN to localize the discriminative region, which is guided by the saliency information extracted by SEN. Faster R-CNN is composed by region proposal network (PRN) and Fast R-CNN \cite{Girshick_2015_ICCV}, both of them share convolutional layers for better performance. 

\begin{comment}
\begin{table*}[!ht]
  \centering
  \caption{Architectures of SEN, RPN and Fast R-CNN in our approach. The layers in blue color are same in three networks. }
  \label{nets}
  \begin{tabular} {|c|c|}
    \hline
    Networks & layers \\
    \hline
    SEN &  {\begin{tabular}{c} {\color{blue}{conv1\_1, conv1\_2, pool1, conv2\_1, conv2\_2, pool2, conv3\_1, conv3\_2, conv3\_3, pool3, conv4\_1, conv4\_2,}} \\ {\color{blue}{ conv4\_3, pool4, conv5\_1 conv5\_2, conv5\_3,}} CAM\_conv, CAM\_pool, CAM\_fc\_cub \end{tabular} }\\
    \hline
    RPN &  {\begin{tabular}{c} {\color{blue}{conv1\_1, conv1\_2, pool1, conv2\_1, conv2\_2, pool2, conv3\_1, conv3\_2, conv3\_3, pool3, conv4\_1, conv4\_2,}} \\ {\color{blue}{ conv4\_3, pool4, conv5\_1 conv5\_2, conv5\_3,}} conv\_proposal1, proposal\_cls\_score, proposal\_bbox\_pred \end{tabular} }\\
    \hline
    Fast R-CNN &  {\begin{tabular}{c} {\color{blue}{conv1\_1, conv1\_2, pool1, conv2\_1, conv2\_2, pool2, conv3\_1, conv3\_2, conv3\_3, pool3, conv4\_1, conv4\_2,}} \\ {\color{blue}{ conv4\_3, pool4, conv5\_1 conv5\_2, conv5\_3,}} roi\_pool5, fc6, fc7, cls\_score, bbox\_pred \end{tabular} }\\
    \hline
  \end{tabular}
\end{table*}
\end{comment}
Instead of using time-consuming bottom-up process such as selective search \cite{uijlings2013selective}, RPN is adopted to quickly generate region proposals of images by sliding a small network over the feature map of last shared conolutional layer. At each sliding-window location, $k$ region proposals are simultaneously predicted, and they are parameterized relative to $k$ anchors. We apply $9$ anchors with $3$ scales and $3$ aspect ratios as Faster R-CNN. For training RPN, a binary class label of being an object or not is assigned to each anchor, which depends on the Intersection-over-Union (IoU) \cite{everingham2015pascal} overlap with a ground truth bounding box of object. But in our approach, we compute the IoU overlap with the bounding box of discriminative region generated by SEN rather than the ground truth bounding box of object, which avoids using the labor-consuming object and parts annotations. And the loss function for an image is defined as:
\begin{gather}
L(\{p_i\},\{t_i\})=\frac{1}{N_{cls}} \sum_i L_{cls}(p_i,p_i^*)\nonumber \\ + \lambda \frac{1}{N_{reg}} \sum_i {p_i^*L_{reg}(t_i,t_i^*)}
\end{gather}
where $i$ denotes the index of an anchor in a mini-batch, $p_i$ denotes the predicted probability of anchor $i$ being a discriminative region, $p_i^*$ denotes the label of being a discriminative region of object or not depending on the bounding box $t_i^*$ generated by SEN , $t_i$ is the predicted bounding box of discriminative region, $L_{cls}$ is the classification loss defined by log loss, and $L_{reg}$ is the regression loss defined by the robust loss function (smooth $L_1$) \cite{Girshick_2015_ICCV}. 

For the localization network, Fast R-CNN \cite{Girshick_2015_ICCV} is adopted. In Fast R-CNN, a region of interest (RoI) pooling layer is employed to extract a fixed-length feature vector from feature map for each region proposal generated by RPN. And each feature vector passes forward for two outputs: one is predicted subcategory and the other is predicted bounding box of discriminative region. 
Through Faster R-CNN, we obtain the discriminative region and subcategory of each image simultaneously, accelerating classification speed.

\subsection{Saliency-guided localization learning}
The saliency-guided localization learning schedules the training process of SEN and Faster R-CNN to make full use of their advantages: (1) SEN learns the saliency information of image to tell which region is important and discriminative for classification, and saliency information guides the training of Faster R-CNN, and (2) RPN in Faster R-CNN generates region proposals that relevant to the discriminative regions of images, which accelerates the process of region proposal rather than using bottom-up process as selective search \cite{uijlings2013selective}. Considering that training RPN needs bounding boxes of discriminative regions provided by SEN, and Fast R-CNN utilizes the proposals generated by RPN, we adopt the strategy of sharing convolutional weights between SEN and Faster R-CNN to promote the localization learning.

%Separately training SEN and Faster R-CNN can achieve the final goal, but ignores that RPN needs the bounding box generated by SEN to guide its training. Therefore, we consider sharing convolutional weights between SEN and RPN for transferring the learned knowledge of discriminative localization. 
First, we train the SEN. This network is first pre-trained on the ImageNet 1K dataset \cite{imagenet_cvpr09}, and then fine-tuned on the fine-grained image classification dataset, such as CUB-200-2011 \cite{wah2011caltech} in our experiment. And then, we train the PRN. Its initial weights of convolutional layers are cloned from SEN. Instead of fixing the shared convolutional layers, all layers are fine-tuned in the training phase. 
Besides, we train RPN and Fast R-CNN follows the strategy in Ren et al. \cite{ren2015faster}.  
 
\section{Experiments}
\subsection{Dataset and evaluation metrics}
We conduct experiments on the widely-used CUB-200-2011 \cite{wah2011caltech} dataset in fine-grained image classification. Our proposed Saliency-guided Faster R-CNN approach is compared with 18 state-of-the-art methods to verify its effectiveness. 

\textbf{CUB-200-2011} \cite{wah2011caltech} is the most widely-used dataset in fine-grained image classification task, which contains 11788 images of 200 subcategories belonging to bird, 5994 images for training and 5794 images for testing. And each image has detailed annotations: a image-level subcategory label, a bounding box of object, and 15 part locations. In our experiments, only image-level subcategory label is used to train the networks.

\textbf{Accuracy} is adopted to comprehensively evaluate the classification performances of our Saliency-guided Faster R-CNN approach and compared methods, which is widely used in fine-grained image classification \cite{zhang2016weakly,zhang2016picking,zhang2014part}, and its definition is as follow:
\begin{gather}
Accuracy = \frac{R_a}{R}
\end{gather} 
where $R$ denotes the number of images in testing set, and $R_a$ denotes the number of images that are correctly classified. 

\textbf{Intersection-over-Union (IoU)} \cite{everingham2015pascal} is adopted to evaluate whether the predicted bounding box of discriminative region is a correct localization, and its formula is defined as: 
\begin{gather}
IoU = \frac{area(B_p \cap B_{gt})}{area(B_p \cup B_{gt})}
\end{gather}
where $B_p$ denotes the predicted bounding box of discriminative region, $B_{gt}$ denotes the ground truth bounding box of object, $B_P \cap B_{gt}$ denotes the intersection of the predicted
and ground truth bounding boxes, and $B_p \cup B_{gt}$ denotes their union. We consider the predicted bounding box of discriminative region is correctly localized, if the IoU exceeds 0.5.
\subsection{Details of the networks}
Our Saliency-guided Faster R-CNN approach consists of three networks: saliency extraction network (SEN), region proposal network (RPN) and Fast R-CNN. They are all based on 16-layer VGGNet \cite{simonyan2014very}, which is widely used in image classification task. The basic CNN can be replaced with the other CNN. 
%The detailed architecture information of the three networks is shown in Table \ref{nets}, where the layers in blue color are same in three networks, and the layers of relu and dropout are not shown for clarity.
SEN extracts the saliency information of images to provide bounding boxes needed by Faster R-CNN. For VGGNet in SEN, we remove the layers after conv$5\_3$ and add a convolutional layer of size $3 \times 3$, stride $1$, pad $1$ with $1024$ neurons, which is followed by a global average pooling layer and a softmax layer \cite{zhou2016cvpr}. We adopt the object-level attention of Xiao et al. \cite{xiao2015application} to select relevant image patches for data extension. And then we utilize the extended data to fine-tune SEN for learning discriminative features. The number of neurons in softmax layer is set as the number of subcategories in the dataset. Faster R-CNN shares the weights of layers before conv$5\_3$ with SEN for better discriminative localization as well as classification performance. The architecture of Fast R-CNN is the same with VGGNet except that pool5 layer is replaced by a RoI pooling layer, and has two outputs: one is predicted subcategory and the other is predicted bounding box of discriminative region.

At training phase, for SEN, we initialize the weights with the network pre-trained on the ImageNet 1K dataset, and then use SGD with a minibatch size of 20. We use a weight decay of 0.0005 with a momentum of 0.9 and set the initial learning rate to 0.001. The learning rate is divided by 10 every 5K iterations. We terminate training at 35K iterations. For Faster RCNN, we initialize the weights with the SEN, and then start SGD with a minibatch size of 128. We use a weight decay of 0.0005 with a momentum of 0.9 and set the initial learning rate to 0.001. We divide the learning
rate by 10 at 30K iterations, and terminate training at 50K iterations.

\subsection{Comparisons with state-of-the-art methods}
This subsection presents the experimental results and analyses of our Saliency-guided Faster R-CNN approach as well as the state-of-the-art methods on the widely-used CUB-200-2011 \cite{wah2011caltech} dataset. We verify the effectiveness of our approach from accuracy and efficiency of classification. 

\begin{table*}[!ht]
  \centering
  \caption{Comparisons with State-of-the-art Methods on CUB-200-2011 dataset.}
  \label{cub}
  \scalebox{1.07}{
  \begin{tabular} {|c|c|c|c|c|c|c|c|}
    \hline
    \multirow {2}{*}{Method} & \multicolumn{2}{c|}{Train Annotation} & \multicolumn{2}{c|}{Test Annotation} & \multirow {2}{*}{Accuracy (\%)} & \multirow {2}{*}{Net}\\
    \cline{2-5}
    &Object & Parts & Object & Parts & &\\
    \hline
    \textbf{Our Saliency-guided Faster R-CNN Approach} & & & & & {\textbf{85.14}} & VGGNet \\
    \hline
    TSC \cite{he2017spatial} & & & & & 84.69 & VGGNet \\
    FOAF \cite{zhang2016fused} & & & & & 84.63 & VGGNet \\
    PD \cite{zhang2016picking}& & & & & 84.54 & VGGNet \\
    %STN \cite{jaderberg2015spatial}& & & & & 84.10 & GoogleNet \\
    Bilinear-CNN \cite{lin2015bilinear}&  & &  & & 84.10 & VGGNet\&VGG-M \\
    %PD+FC-CNN \cite{zhang2016picking} & & & & & 82.60 & VGGNet \\
    %Multi-grained \cite{wang2015multiple} & & & & & 81.70 & VGGNet \\
    NAC \cite{simon2015neural} & & & & & 81.01 & VGGNet \\
    PIR \cite{zhang2016weakly}& & & & & 79.34 & VGGNet \\
    TL Atten \cite{xiao2015application} & & & & & 77.90 & VGGNet \\
    MIL \cite{xu2017friend} & & & & & 77.40 & VGGNet \\
    %VGG-BGLm \cite{zhou2016fine} & & & & & 75.90 & VGGNet \\
    %Dense Graph Mining \cite{zhang2016detecting} & & & & & 60.19 &  \\
    \hline
    %Coarse-to-Fine \cite{yao2016coarse}& $\surd$ & & & & 82.50 & VGGNet \\
    %FOAF \cite{zhang2016fused} & $\surd$ & & $\surd$ & & 86.34 & VGGNet \\
    %Bilinear-CNN \cite{lin2015bilinear}& $\surd$ & & $\surd$ & & 85.10 & VGGNet\&VGG-M \\
    Coarse-to-Fine \cite{yao2016coarse}& $\surd$ & & $\surd$ & & 82.90 & VGGNet \\
    PG Alignment \cite{krause2015fine} & $\surd$ & & $\surd$ & & 82.80 & VGGNet \\
    VGG-BGLm \cite{zhou2016fine} & $\surd$ & & $\surd$ & & 80.40 & VGGNet \\
    %Triplet-A (64) \cite{cui2015fine} & $\surd$ & & $\surd$ & & 80.70 & GoogleNet \\
    %Triplet-M (64) \cite{cui2015fine} & $\surd$ & & $\surd$ & & 79.30 & GoogleNet \\
    \hline
    Webly-supervised \cite{xu2016webly} & $\surd$ & $\surd$ &  &  & 78.60 & AlexNet \\
    PN-CNN \cite{branson2014bird} & $\surd$ & $\surd$ &  &  & 75.70 & AlexNet \\
    Part-based R-CNN \cite{zhang2014part} & $\surd$ & $\surd$ &  & & 73.50 & AlexNet \\
    SPDA-CNN \cite{zhang2016spda} & $\surd$ & $\surd$ & $\surd$ &  & 85.14 & VGGNet \\
    Deep LAC \cite{lin2015deep} & $\surd$ & $\surd$ & $\surd$ &  & 84.10 & AlexNet \\
    %SPDA-CNN \cite{zhang2016spda} & $\surd$ & $\surd$ & $\surd$ &  & 81.01 & AlexNet \\
    PS-CNN \cite{huang2016part}& $\surd$ & $\surd$ & $\surd$ &  & 76.20 & AlexNet \\
    PN-CNN \cite{branson2014bird} & $\surd$ & $\surd$ & $\surd$ & $\surd$ & 85.40 & AlexNet  \\
    %Part-based R-CNN \cite{zhang2014part} & $\surd$ & $\surd$ & $\surd$ & $\surd$ & 76.37 & AlexNet \\
    POOF \cite{berg2013poof} & $\surd$ & $\surd$ & $\surd$ & $\surd$ & 73.30 &  \\
    %GPP \cite{xie2013hierarchical} & $\surd$ & $\surd$ & $\surd$ & $\surd$ & 66.35 &  \\
    \hline
  \end{tabular}}
\end{table*}

\subsubsection{Accuracy of classification}
Table \ref{cub} shows the comparison results on CUB-200-2011 dataset at the aspect of classification accuracy. Object, parts annotations and CNN used in these methods are listed for fair comparison. 
%The compared methods are grouped into three sets: (1) The first set contains the methods that neither object nor parts annotations are used in both training or testing phase. (2) The second set contains the methods that only use the object annotations. (3) And The third set contains the methods that both object and parts annotations are used. 
Traditional methods as \cite{berg2013poof} choose SIFT \cite{lowe2004distinctive} as features, even using both object and parts annotations its performance is limited and much lower than our approach. 
Our approach achieves the highest classification accuracy among all methods under the same weakly supervised setting that neither object nor parts annotations are used in training or testing phase, 
and obtains 0.45\% higher accuracy than the best result of TSC \cite{he2017spatial} (85.14\% vs. 84.69\%), which jointly considers two spatial constraints in parts selection. Despite achieving better classification accuracy, our approach is over two order of magnitude faster than TSC, due to the end-to-end network, as shown in Table \ref{cubtime}. The efficiency analysis will be described in Section 3.3.2.
And our approach performs better than the method of Bilinear-CNN \cite{lin2015bilinear}, which combines two different CNNs: VGGNet \cite{simonyan2014very} and VGG-M \cite{chatfield2014return}. Its classification accuracy is 84.10\%, which is lower than our approach by 1.04\%. 
Furthermore, our approach even outperforms these methods using object annotation in both training and testing phase by at least 2.24\%, such as Coarse-to-Fine \cite{yao2016coarse}, PG Alignment \cite{krause2015fine} and VGG-BGLm \cite{zhou2016fine}. 
Moreover, our approach outperforms these methods that use both object and parts annotations \cite{zhang2014part,xu2016webly}. 
%Only beaten by PN-CNN \cite{huang2016part}, which uses both object and parts annotations in not only training but also testing phase. Even though, its classification accuracy is only 0.26\% higher than ours. 
Neither object nor parts annotations are used in our Saliency-guided Faster R-CNN approach, which leads fine-grained image classification to practical application. Besides, end-to-end network in our approach simultaneously localizes discriminative region and encodes discriminative feature for each image, and discriminative localization promotes the classification performance.
 
\begin{table*}[!ht]
  \centering
  \caption{Comparison of average classification speed (frames per second) with state-of-the-art methods on CUB-200-2011 dataset. The results are obtained on the computer with NVIDIA TITAN X @1417MHZ and Intel Core i7-6900K @3.2GHZ.}
  \label{cubtime}
  \begin{tabular} {|p{7.4cm}<{\centering}|p{4cm}<{\centering}|p{4cm}<{\centering}|}
    \hline
    Methods & Testing Speed (fps)  & Net\\
    \hline
    \textbf{Our Saliency-guided Faster R-CNN Approach} & \textbf{10.07} &  VGGNet \\
    Bilinear-CNN \cite{lin2015bilinear}& 4.52 &  VGGNet\&VGG-M \\
    TSC \cite{he2017spatial} & 0.34 &  VGGNet \\
    TL Atten \cite{xiao2015application} & 0.25 &VGGNet \\
    NAC \cite{simon2015neural} & 0.10  &VGGNet \\
    %PG Alignment \cite{krause2015fine} & & VGGNet  \\
    \hline
    \textbf{Our Saliency-guided Faster R-CNN Approach}  & \textbf{17.09} & AlexNet \\
    PS-CNN \cite{huang2016part} & 14.30 & AlexNet\\
    \hline
  \end{tabular}
\end{table*}
\begin{table*}[!ht]
  \centering
  \caption{Classification and localization Accuracies.}
  \label{component}
  \begin{tabular} {|p{7.4cm}<{\centering}|p{4cm}<{\centering}|p{4cm}<{\centering}|}
    \hline
    Methods & Classification Accuracy(\%) & Localization  Accuracy(\%)  \\
    \hline
    \textbf{Our Saliency-guided Faster R-CNN Approach} & \textbf{85.14} & \textbf{46.05}\\
    %\hline
    %Fast R-CNN(stage1) & 84.76 & 45.84 \\
    SEN & 77.50 & 44.93\\
    %CAM(stage1) & 76.26 & 34.29\\ 
    \hline
  \end{tabular}
\end{table*}
\begin{figure*}[!ht]
    \begin{center}\includegraphics[width=0.95\linewidth]{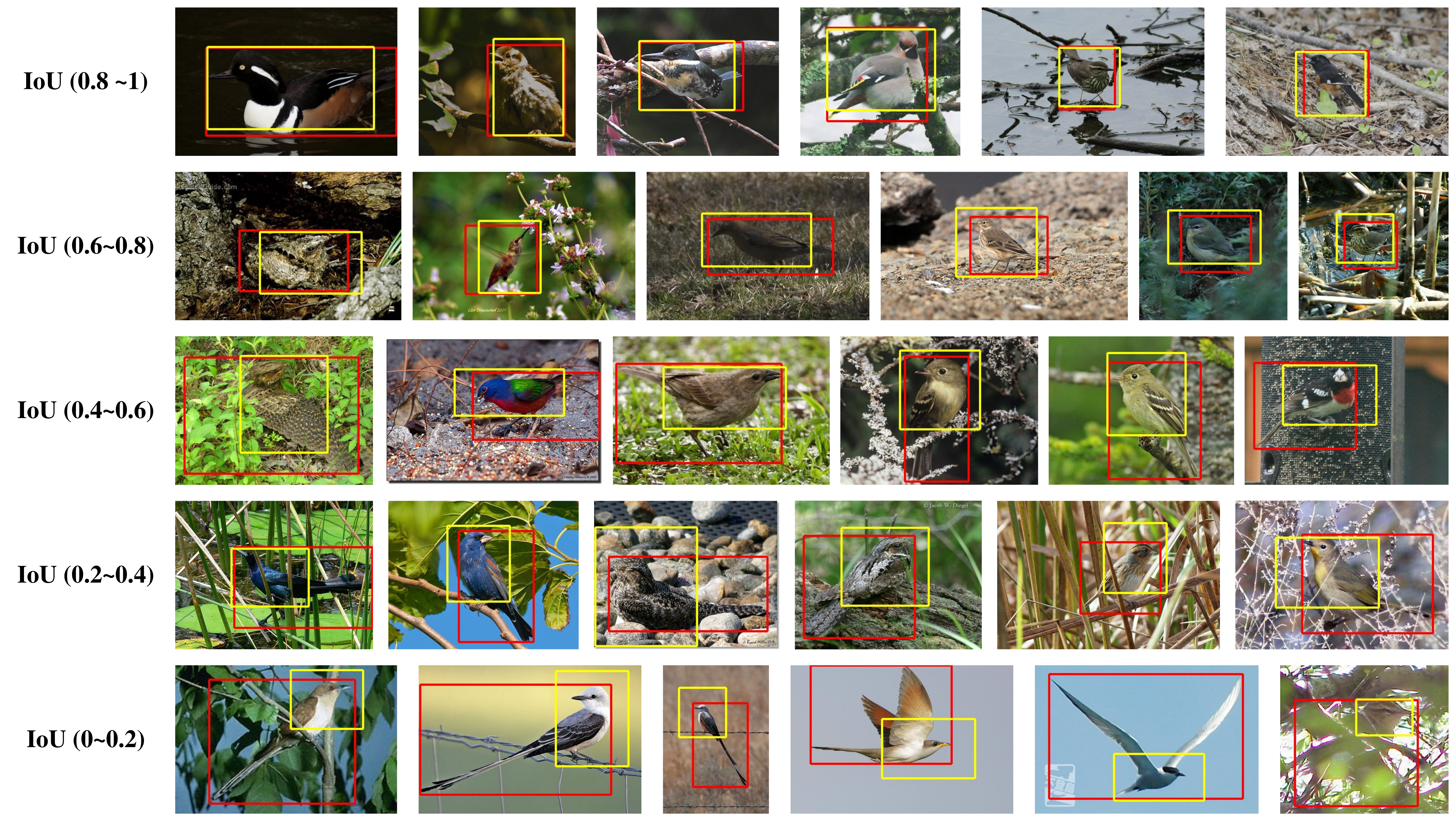}
    \caption{Samples of predicted bounding boxes of discriminative regions (yellow rectangles) and ground truth bounding boxes of objects (red rectangles) at different ranges of IoU on CUB-200-2011 dataset. }
    \label{boundingbox}
    \end{center}
\end{figure*}

\subsubsection{Efficiency of classification}
Experimental results at the aspect of efficiency on CUB-200-2011 dataset is presented in Table \ref{cubtime}. We get the testing speed on the computer with NVIDIA TITAN X @1417MHZ and Intel Core i7-6900K @3.2GHZ, and use frames per second (fps) to evaluate the efficiency. 
Comparing with state-of-the-art methods, our Saliency-guided Faster R-CNN approach achieves the best performance on not only the classification accuracy but also the efficiency. 
%Only the testing time of Bilinear-CNN is close to ours, which is an end-to-end network combing two CNNs. Even the testing time is at the same magnitude, but Bilinear-CNN can not localize the discriminative regions and its classification accuracy is 1.04\% lower than ours.
We split state-of-the-art methods into two groups by the basic CNNs used in their methods: VGGNet \cite{simonyan2014very} and AlexNet \cite{krizhevsky2012imagenet}. Results of these methods in first group are obtained by their authors' source codes. 
Comparing with these methods, our approach improves about 123\% than Bilinear-CNN at the aspect of classification speed (10.07 fps vs. 4.52 fps). Besides, our classification accuracy is also 1.04\% higher than Bilinear-CNN.  
Even more, our approach is over two orders of magnitude faster than these methods with two separated stages of localization and encoding. 
When utilizing AlexNet as the basic network, our approach is still faster than PS-CNN \cite{huang2016part} and improves about 19.51\%, which also utilizes AlexNet. And when applying AlexNet as basic CNN in our approach, the classification accuracy is 73.58\%. It is noted that neither object nor parts annotations are used in our approach, while all used in PS-CNN.
The classification speed of PS-CNN \cite{huang2016part} is reported as 20 fps in their paper. They provide a reference that a single CaffeNet \cite{jia2014caffe} runs at 50 fps under their experimental setting (NVIDIA Tesla K80). In our experiments, a single CaffeNet runs at 35.75 fps, so we calculate the speed of PS-CNN in the same experimental setting with ours as 20*35.75/50=14.30 fps. 
%To our best knowledge, PS-CNN is the fastest fine-grained image classification method with discriminative localization. 
Our approach avoids the time-consuming classification process by the design of end-to-end network, and achieves the best classification performance by the mutual promotion between localization and classification. This leads the fine-grained image classification to practical application. 

\subsection{Effectiveness of discriminative localization}

Saliency-guided localization learning is proposed to train SEN and Faster R-CNN for improving the localization and classification performance simultaneously. Since we devote to localizing the discriminative region which is generally located at the object, we adopt the IoU overlap between discriminative region and ground truth bounding box of object to evaluate the correctness of localization. We consider a bounding box of discriminative region to be correctly predicted if IoU with ground truth bounding box of object is larger than 0.5. The accuracy of localization is shown in Table \ref{component}. 

Our Saliency-guided Faster R-CNN approach achieves the accuracy of 46.05\%. Considering that neither object nor parts annotations are used, it is a promising result. And comparing with ``SEN'' which means directly using SEN to generate bounding box, our approach achieves improvements both in classification and localization, which verifies the effectiveness of our saliency-guided localization learning approach.
\begin{table*}[!ht]
  \centering
  \caption{\emph{PCL} for each part of object in the CUB-200-2011 testing set.}
  \label{parts}
  \begin{tabular} {|p{1.47cm}<{\centering}|p{1.47cm}<{\centering}|p{1.47cm}<{\centering}|p{1.47cm}<{\centering}|p{1.47cm}<{\centering}|p{1.47cm}<{\centering}|p{1.47cm}<{\centering}|p{1.47cm}<{\centering}|p{1.47cm}<{\centering}|}
    \hline
    Parts & back & beak & belly & breast & crown & forehead & left eye & left leg\\
    \hline
    PCL (\%) & 96.33 & 96.49 & 94.00 & 95.29 & 97.38 & 97.07 & 97.49 & 89.92 \\
    \hline
    Parts & left wing & nape & right eye & right leg & right wing & tail & throat & \textbf{average}\\
    \hline
    PCL (\%)& 92.60 & 96.60 & 96.79 & 91.85 & 97.00 & 85.03 & 96.38 & \textbf{94.68}\\
    \hline
  \end{tabular}
\end{table*}
\begin{figure}[!ht]
    \begin{center}\includegraphics[width=0.9\linewidth]{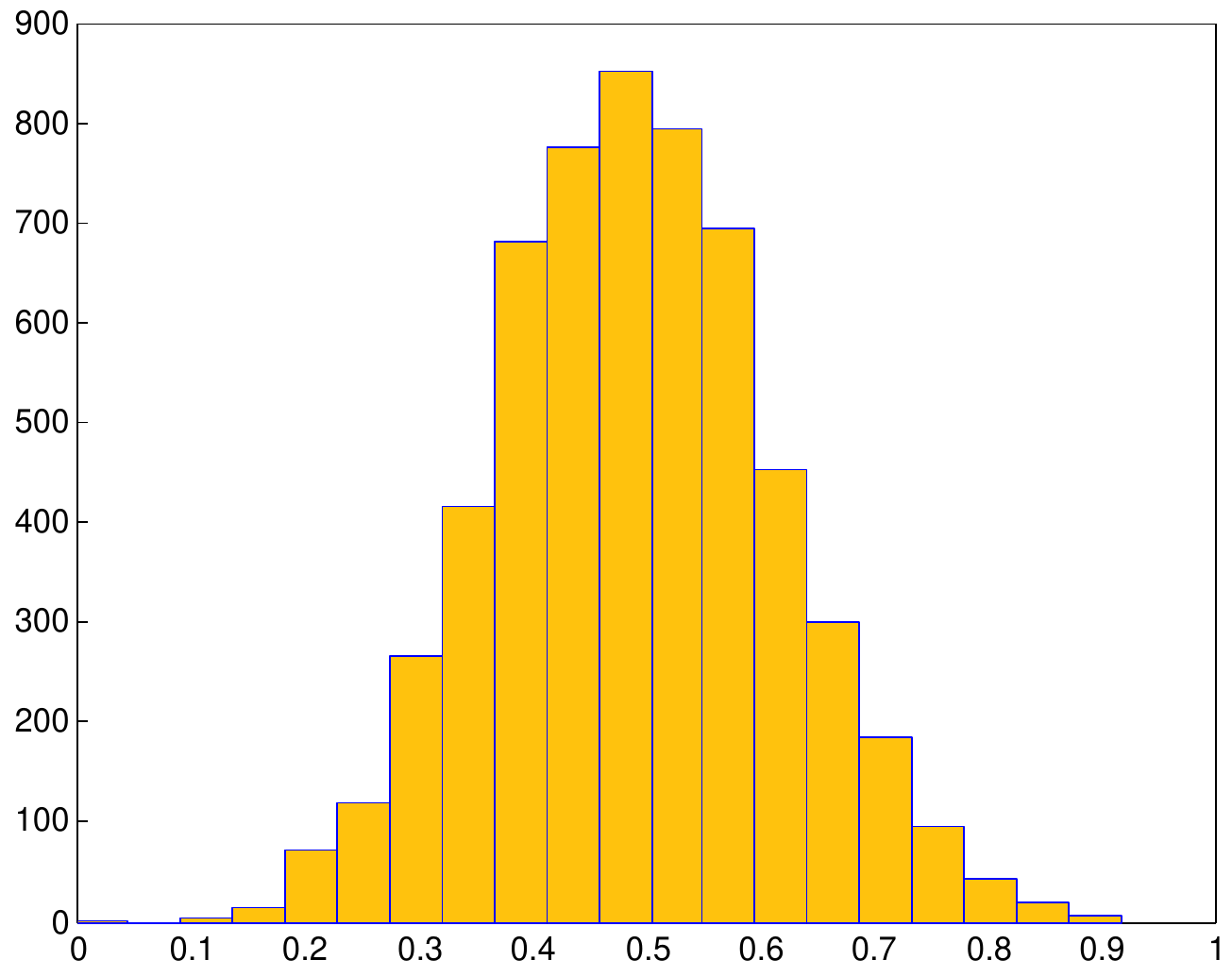}
    \caption{IoU histogram. Abscissa denotes the IoU overlap between predicted bounding box of discriminative region and ground truth bounding box of object. And ordinate means the number of testing images that have the IoU overlap at the range. }
    \label{ioucurve}
    \end{center}
\end{figure} 
We show some samples of predicted bounding boxes of discriminative regions and ground truth bounding boxes of objects at different ranges of IoU (e.g. 0$\sim$0.2, 0.2$\sim$0.4, 0.4$\sim$0.6, 0.6$\sim$0.8, 0.8$\sim$1) on CUB-200-2011 dataset, as Figure \ref{boundingbox}. We have some predicted bounding boxes whose IoUs with ground truth bounding boxes of objects are lower than 0.5. But these predicted bounding boxes contain discriminative regions of objects, such as heads and bodies. 
It verifies the effectiveness of our approach in localizing discriminative region of object for achieving better classification performance. 
Figure \ref{ioucurve} shows the histogram of IoU. We can observe that most testing images lie in the range of 0.4$\sim$1. 
To further verify the effectiveness of discriminative localization in our approach, results are given in terms of the  Percentage of Correctly Localization (PCL) in Table \ref{parts}, estimating whether the predicted bounding box contains the parts of object or not. CUB-200-2011 dataset provides 15 part locations, which denote the pixel locations of centers of parts. We consider our predicted bounding box contain a part if the part location lies in the area of the predicted bounding box. Table \ref{parts}  shows that about average 94.68\% of the parts located in our predicted bounding boxes. It shows that our discriminative localization can detect the distinguishing information of objects to promote classification performance.

\subsection{Analysis of misclassification} 
\begin{figure}[!t]
    \begin{center}\includegraphics[width=0.95\linewidth]{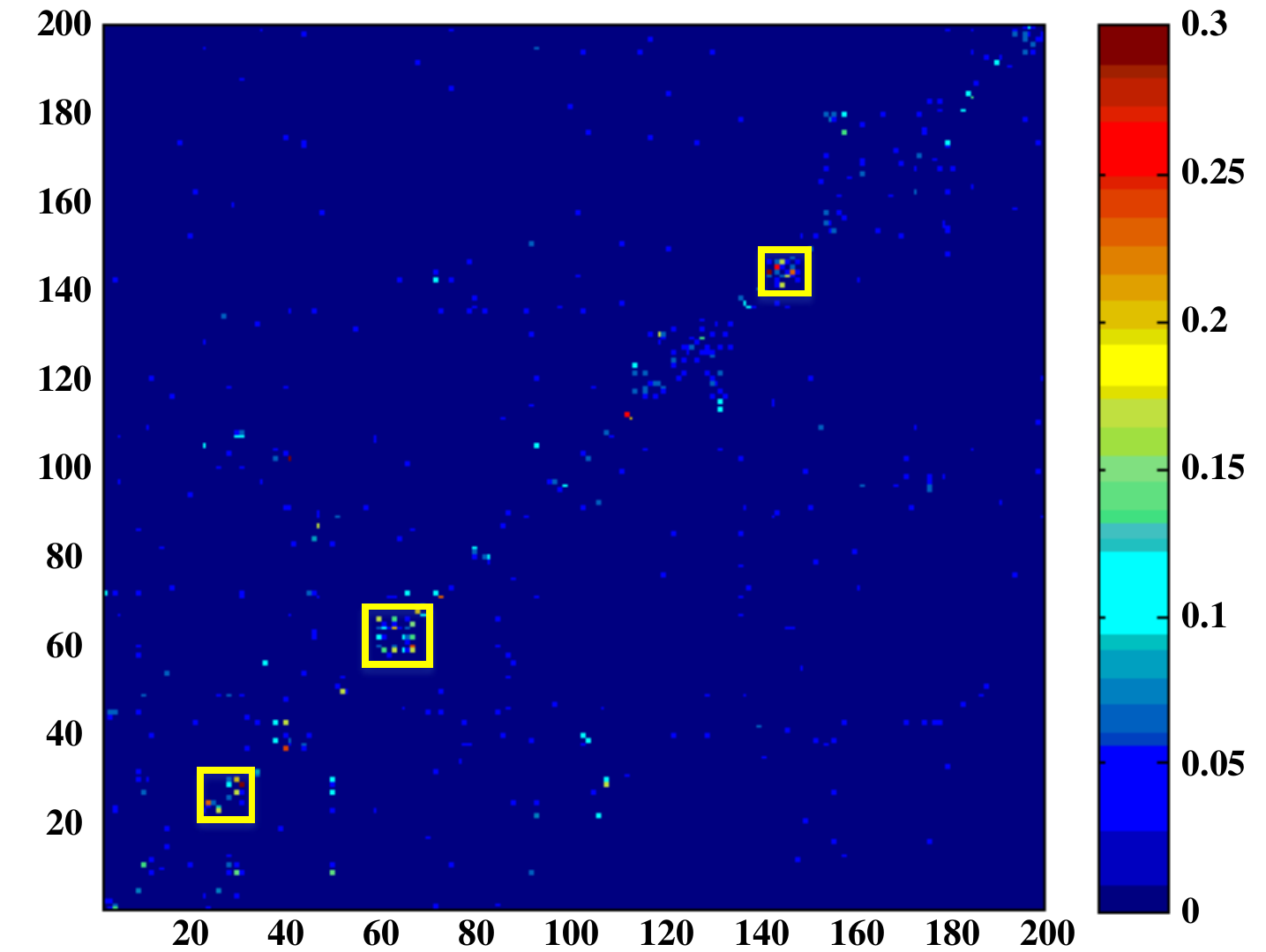}
    \caption{Classification confusion matrix on CUB-200-2011 dataset with 200 subcategories. The yellow rectangles show the sets of subcategories with the higher probability of misclassification. }
    \label{confusion}
    \end{center}
\end{figure}

\begin{figure}[!t]
    \begin{center}\includegraphics[width=1\linewidth]{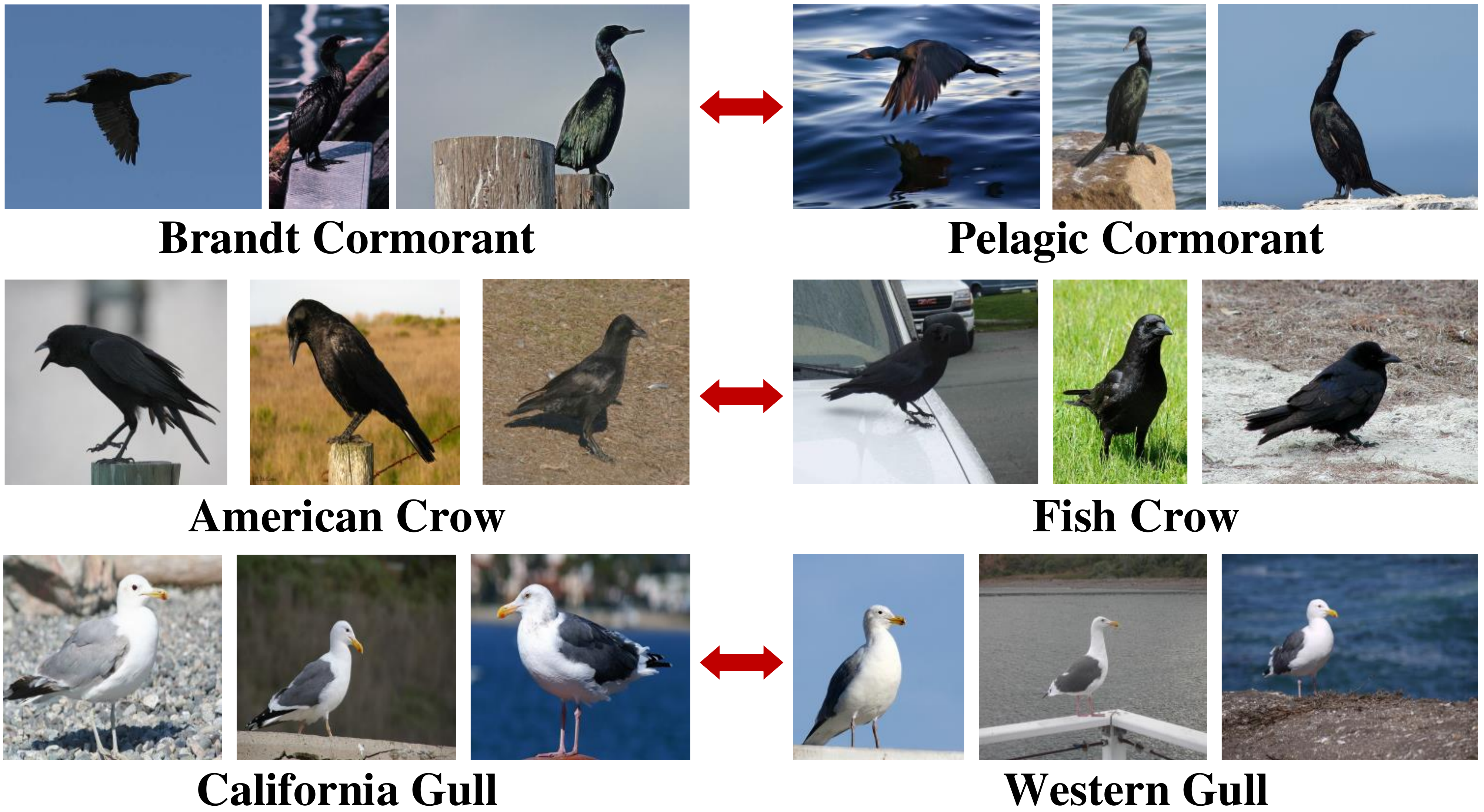}
    \caption{Examples of the most confused subcategory pairs. One subcategory is mostly confidently classified as the other in the same row when in the testing phase.}
    \label{confusionexamples}
    \end{center}
\end{figure}

Figure \ref{confusion} shows the classification confusion matrix for our approach, where coordinate axes denote subcategories and different colors denote different probabilities of misclassification. The yellow rectangles show the sets of subcategories with the higher probability of misclassification. We can observe that these sets of subcategories locate near the diagonal of the confusion matrix, which means that these misclassification subcategories generally belong to the same genus with small variance. The small variance is not easy to measure from the image, which leads the high challenge of fine-grained image classification. 
Figure \ref{confusionexamples} shows some examples of the most probably confused subcategory pairs. One subcategory is most confidently classified as the other in the same row. The subcategories in the same row look almost the same, and belong to the same genus. For example, ``Brandt Cormorant'' and ``Pelagic Cormorant'' look the same in the appearance, both of them have the same attributes of black feather and long neck, and belong to the genus of ``Phalacrocorax''.  
It is extremely difficult for us to distinguish between them.
%Besides, we note that CUB-200-2011 dataset has an estimated 4.4\% label noise hence some of these errors may be incorrect \cite{van2015building}.


\subsection{Comparison with baselines}
Our Saliency-guided Faster R-CNN approach is based on Faster-RCNN \cite{ren2015faster}, and adopts VGGNet \cite{simonyan2014very} as the basic model. To verify the effectiveness of our approach, we present the results of our approach as well as the baselines in Table \ref{faster}. ``VGGNet'' denotes the result of directly using fine-tuned VGGNet, and ``Faster R-CNN (gt)'' denotes the result of directly adopting Faster R-CNN with ground truth bounding box to guide training phase. Our approach achieves the best performance even without using object or parts annotations. We adopt VGGNet as the basic model in our approach, but its classification accuracy is only 70.42\%, which is much lower than ours. It shows that the discriminative localization has promoting effect to classification. With discriminative localization, we find the most important regions of images for classification, which contains the key variance from other subcategories. Comparing with ``Faster R-CNN (gt)'', our approach also achieves better performance. It is an interesting phenomenon that worth thinking about. From the last row in Figure \ref{boundingbox}, we observe that not all the areas in the ground truth bounding boxes are helpful for classification. Some ground truth bounding boxes contain large area of background noise that has less useful information and even leads to misclassification. So discriminative localization is significantly helpful for achieving better classification performance. And comparison with ``Ours (without shared conv layers)'' verifies the effectiveness of our saliency-guided localization learning represented in Section 2.2, which promotes not only discriminative localization but also classification.

\begin{comment}
\begin{table}[!ht]
  \centering
  \caption{Precision of the bounding box predicted by our Saliency-guided Faster R-CNN approach on CUB-200-2011 dataset. The precision is defined by the proportion of Intersection-over-Union (IoU) overlap with ground truth bounding box at least 0.5.}
  \label{iou}
  \begin{tabular} {|c|c|}
    \hline
    Methods & Accuracy (\%) \\
    \hline
    {\begin{tabular}{c} \textbf{Our Saliency-guided Faster } \\ \textbf{R-CNN Approach} \end{tabular} } & \textbf{46.05} \\
    \hline
    Fast R-CNN(stage1) & 45.84 \\
    CAM-stage2 & 44.93 \\
    %RPN-stage2 &  \\
    %RPN-stage1 &  \\
    CAM-stage1 & 34.29 \\
    \hline
  \end{tabular}
\end{table}
\end{comment}

\begin{table}[!t]
  \centering
  \caption{Comparison with baselines. }
  \label{faster}
  \begin{tabular} {|c|c|}
    \hline
    Methods & Accuracy (\%) \\
    \hline
    {\begin{tabular}{c} \textbf{Our Saliency-guided Faster } \\ \textbf{R-CNN Approach} \end{tabular} } & \textbf{85.14} \\
    \hline
    Ours (without shared conv layers) & 83.95 \\
    Faster R-CNN (gt) & 82.46 \\
    %Faster R-CNN (ori) & 82.22 \\
    VGGNet & 70.42 \\
    %CAM(bbox) & \\
    \hline
  \end{tabular}
\end{table}
\section{Conclusion}
In this paper, discriminative localization approach via saliency-guided Faster R-CNN has been proposed for weakly supervised fine-grained image classification. 
We first propose saliency-guided localization learning approach to localize discriminative region automatically for each image, which uses neither object nor parts annotations to avoid using labor-consuming annotations. 
And then an end-to-end network based on Faster R-CNN with guide of saliency information is proposed to simultaneously localize discriminative region and encode discriminative features, which not only achieves a notable classification performance but also accelerates classification speed. And combining them, we simultaneously accelerate classification speed and eliminate dependence on object and parts annotations.  
Comprehensive experimental results show our Saliency-guided Faster R-CNN approach is more effective and efficient compared with state-of-the-art methods on the widely-used CUB-200-2011 dataset.  

The future works lie in two aspects: First, we will focus on learning better discriminative localization via exploiting the effectiveness of fully convolutional networks. Second, we will also attempt to localize several discriminative regions with different semantic meanings simultaneously, such as the head or body of bird, to improve fine-grained image classification performance. 

\section{ Acknowledgments}
This work was supported by National Natural Science Foundation of China under Grants 61371128 and 61532005.

\bibliographystyle{ACM-Reference-Format}
\balance
\bibliography{sigproc}

\end{document}